\documentclass[letterpaper]{article} 

\ifdefined\aaaianonymous
    \usepackage[submission]{aaai2026}  
\else
    \usepackage{aaai2026}              
\fi

\usepackage{times}  
\usepackage{helvet}  
\usepackage{courier}  
\usepackage[hyphens]{url}  
\usepackage{graphicx} 
\usepackage{natbib}  
\usepackage{caption} 
\usepackage{algorithm}
\usepackage{algorithmic}

\usepackage{newfloat}
\usepackage{listings}
\DeclareCaptionStyle{ruled}{labelfont=normalfont,labelsep=colon,strut=off} 
\floatstyle{ruled}
\newfloat{listing}{tb}{lst}{}
\floatname{listing}{Listing}

\usepackage{amsmath}
\usepackage{cleveref}
\usepackage{subfloat}
\usepackage{booktabs}
\usepackage{amsfonts}
\usepackage{bm}
\usepackage{multirow}
\usepackage{txfonts}
\usepackage{pifont}
\usepackage{makecell}
\usepackage{enumitem}
\usepackage{threeparttable}
\RequirePackage[format=plain,labelformat=simple,labelsep=period,font=small,compatibility=false]{caption}
\RequirePackage[font=footnotesize,skip=3pt,subrefformat=parens]{subcaption}

\ifdefined\aaaianonymous
    \title{Region-to-Region: Enhancing Generative Image Harmonization with \par Adaptive Regional Injection}
\else
    \title{Region-to-Region: Enhancing Generative Image Harmonization with \par Adaptive Regional Injection}
\fi

\author {
    Zhiqiu Zhang\textsuperscript{\rm 1},
    Dongqi Fan\textsuperscript{\rm 1},
    Mingjie Wang\textsuperscript{\rm 2},
    Qiang Tang\textsuperscript{\rm 3},
    Jian Yang\textsuperscript{\rm 1},
    Zili Yi\textsuperscript{\rm 1}
}
\affiliations {
    \textsuperscript{\rm 1}Nanjing University\\
    \textsuperscript{\rm 2}Zhejiang Sci-Tech University\\
    \textsuperscript{\rm 3}Electronic Arts\\
    zhiqiuzhang@smail.nju.edu.cn
}

\usepackage{bibentry}

\begin{document}

\maketitle

\begin{abstract}
The goal of image harmonization is to adjust the foreground in a composite image to achieve visual consistency with the background.
Recently, latent diffusion model (LDM) are applied for harmonization, achieving remarkable results.
However, LDM-based harmonization faces challenges in detail preservation and limited harmonization ability.
Additionally, current synthetic datasets rely on color transfer, which lacks local variations and fails to capture complex real-world lighting conditions.
To enhance harmonization capabilities, we propose the \textit{Region-to-Region} transformation. By injecting information from appropriate regions into the foreground, this approach preserves original details while achieving image harmonization or, conversely, generating new composite data.
From this perspective, We propose a novel model \textit{R2R}.
Specifically, we design \textit{Clear-VAE} to preserve high-frequency details in the foreground using Adaptive Filter while eliminating disharmonious elements. 
To further enhance harmonization, we introduce the \textit{Harmony Controller} with Mask-aware Adaptive Channel Attention (\textit{MACA}), which dynamically adjusts the foreground based on the channel importance of both foreground and background regions.
To address the limitation of existing datasets, we propose \textit{Random Poisson Blending}, which transfers color and lighting information from a suitable region to the foreground, thereby generating more diverse and challenging synthetic images. 
Using this method, we construct a new synthetic dataset, \textit{RPHarmony}.
Experiments demonstrate the superiority of our method over other methods in both quantitative metrics and visual harmony.
Moreover, our dataset helps the model generate more realistic images in real examples.
Our code, dataset, and model weights have all been released for open access.
\end{abstract}

\ifdefined\aaaianonymous
\else
\begin{links}
    \link{Code}{https://github.com/anonymity-111/Region_to_Region}
\end{links}
\fi

\section{Introduction}

Compositing images with inconsistent conditions creates unrealistic results. Image harmonization refines the foreground to align with the background's color and lighting, ensuring a more natural blend.

Numerous methods have been proposed for the image harmonization task.
Traditionally, this problem has been approached either through image gradient-based techniques~\cite{gradient1,gradient2} for seamless blending or from a statistical perspective~\cite{colorstatistic1,colorstatistic2}.
With the advent of deep learning, learning-based methods trained on large-scale datasets have achieved significant advancements in this field~\cite{cong2020dovenetdeepimageharmonization,ling2021regionawareadaptiveinstancenormalization,chen2023hierarchicaldynamicimageharmonization,xue2022dccfdeepcomprehensiblecolor}.
Recently, diffusion model~\cite{ho2020denoisingdiffusionprobabilisticmodels}, particularly latent diffusion model (LDM)~\cite{rombach2022highresolutionimagesynthesislatent}, have become the mainstream paradigm for image generation.
Some works have applied LDM to address the image harmonization task, leveraging the rich prior knowledge of diffusion models~\cite{li2023imageharmonizationdiffusionmodel,chen2023zero,zhou2024diffharmony,zhou2024diffharmonypp,hachnochi2023crossdomaincompositingpretraineddiffusion}.

While recent advancements have led to significant progress in the qualitative metrics of datasets, state-of-the-art models, including LDM-based methods, still struggle with real composite images. This shortcoming arises from both the model architecture and the limitations of current training datasets.

From a model perspective, applying LDM to image harmonization presents two key challenges. First, the VAE encoding process can result in a loss of fine details, a concern also noted in \cite{yu2024promptfixpromptfixphoto, zhou2024diffharmony, zhou2024diffharmonypp}. While some methods attempt to address this, they often preserve details at the expense of introducing disharmonious elements from the original foreground. Second, the vanilla LDM, designed for general image generation, has inherent limitations in harmonization capabilities. 
Resolving these two challenges will further unleash the potential of LDM in advancing image harmonization.

The limitations of existing datasets also compound this issue. Mainstream datasets, such as iHarmony4~\cite{cong2020dovenetdeepimageharmonization}, are typically generated using the color transfer method~\cite{tsai2017deepimageharmonization}. This technique modifies the foreground of a real image by referencing semantically similar objects from another image. However, this approach primarily applies global adjustments, neglecting local variations in color and lighting~\cite{ren2024relightfulharmonizationlightingawareportrait}. As a result, these synthetic images do not fully represent the complexity of real-world lighting conditions, thereby hindering progress in image harmonization.

To address model and dataset limitations, we propose a \textbf{Region-to-Region} transformation approach. Specifically, transformation is achieved by injecting information from appropriate regions—background, composite foreground, or a specific area from a reference image—into the foreground region. 

Compared to global color-to-color or local pixel-to-pixel adjustments~\cite{CDTNet}, our transformation operates at an intermediate scale such as the region (a group of pixels), balancing global and local information.

From this perspective, we design a novel model, \textbf{R2R}. First, we propose \textbf{Clear-VAE}, an improved VAE that incorporates skip connection with Adaptive Filter in each layer to restore high-frequency details from the composite foreground. During training, we introduce contrastive regulation loss~\cite{hang2022scscoselfconsistentstylecontrastive} to remove disharmonious elements in high-frequency details. Moreover, we introduce the \textbf{Harmony Controller}, a ControlNet-like structure~\cite{zhang2023addingconditionalcontroltexttoimage,anydoor}, to enhance the U-Net for generation. Within each block, we integrate Mask-aware Adaptive Channel Attention \textbf{(MACA)}, which adaptively refines foreground features based on the channel importance of both foreground and background regions, guided by the mask.

Meanwhile, we propose \textbf{Random Poisson Blending}, a data synthesis technique motivated by the Region-to-Region transformation paradigm, for constructing a more realistic dataset. 
Our approach uses Poisson Blending~\cite{gradient1} to directly apply background information from random regions in a reference image to modify the foreground of a real image. 
Using this method, we introduce a new dataset, \textbf{RPHarmony}, consisting of 12,787 training images and 1,422 test images. 
With a more diverse and challenging set of composite images, RPHarmony enables models to achieve more effective image harmonization in real-world scenarios after fine-tuning (Fig.~\ref{fig:real-har}).

Our contributions can be summarized as follows:
\begin{itemize}
\item We introduce Region-to-Region transformation for harmonization.
From this perspective, we introduce R2R, a novel model that enhances Generative Image Harmonization through regional injection.
\item We propose Random Poisson Blending, a Region-to-Region transformation method, to generate composite images and create a new dataset, RPHarmony.
Compared to iHarmony4, our dataset provides more realistic and complex composite images, bridging the gap to real-world scenarios.
\item Our model achieves new state-of-the-art performance on the iHarmony4 dataset, with a 0.28dB improvement in PSNR and over 10\% reduction in MSE. 
Additionally, results on real-world composite images indicate that both our R2R model and RPHarmony dataset significantly contribute to improving the generalization capability of harmonization models.
\end{itemize}

\section{Related Work}
\begin{figure*}[t]
    \centering
    \includegraphics[width=1.0\linewidth]{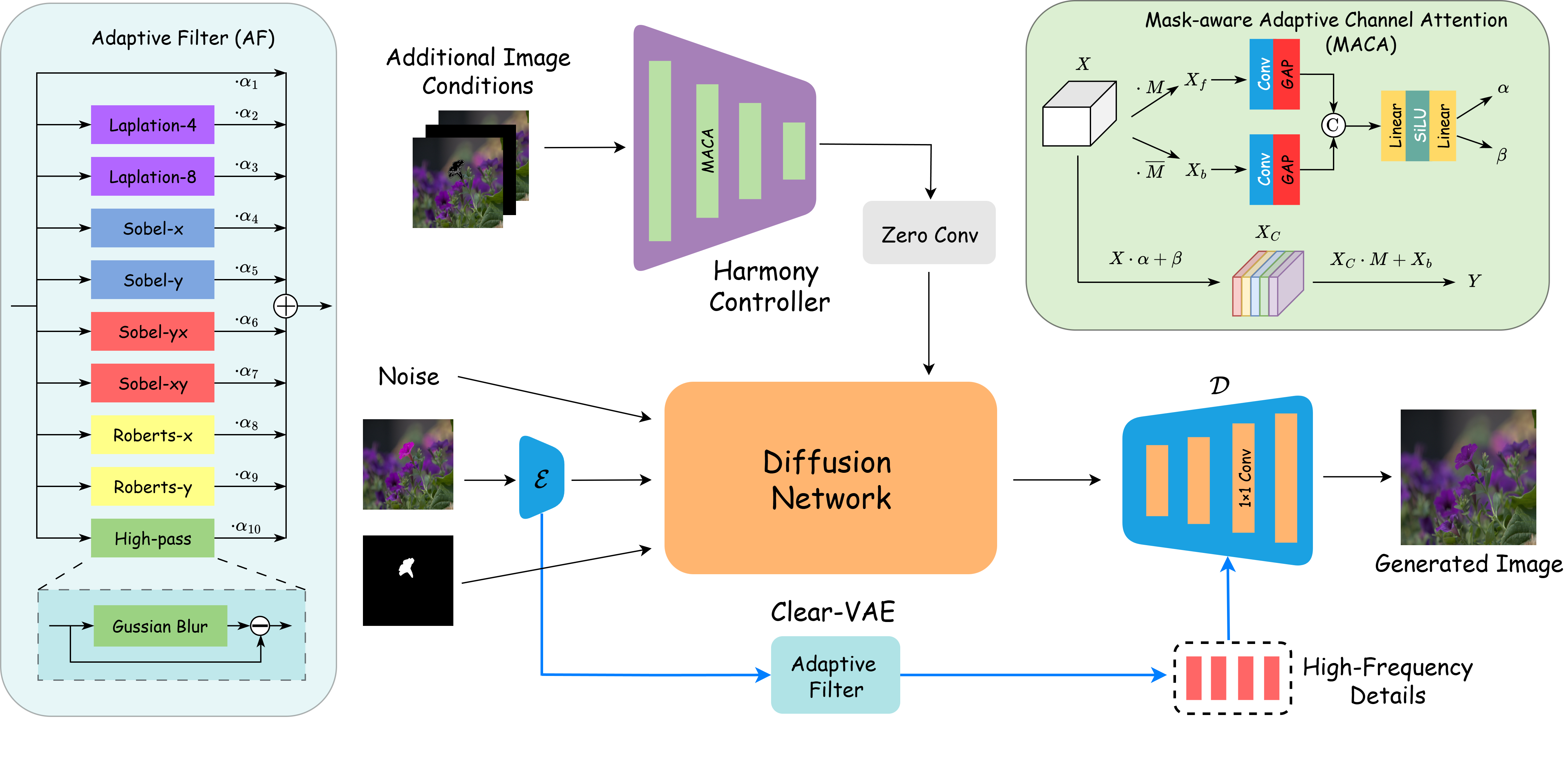}
    \caption{\textbf{Overview of our model structure.} The R2R model consists of three components: the Clear-VAE, which encodes initial composite images and decodes predicted images while using the Adaptive Filter to extract high-frequency details for appearance consistency; the Harmony Controller with MACA, which takes Additional Image Conditions as input to extract and form image guidance conditions; and the Diffusion Network, which performs image harmonization in latent space with the injection from Controller.}
    \label{fig: overview}
\end{figure*}

\subsection{Diffusion Model}

Diffusion model~\cite{ho2020denoisingdiffusionprobabilisticmodels, ddim} has shown strong capabilities in realistic image generation. Latent Diffusion Models (LDMs)~\cite{rombach2022highresolutionimagesynthesislatent} improve efficiency by operating in the latent space. To enable controllable generation, ControlNet~\cite{zhang2023addingconditionalcontroltexttoimage} introduces an auxiliary encoder to inject image conditions into the diffusion process, and has been widely used in image-to-image translation~\cite{anydoor, depth}.

\subsection{Image Harmonization}

Image harmonization aims to adjust the foreground to be visually consistent with the background in terms of color, lighting, and other low-level cues~\cite{niu2024makingimagesrealagain}. Traditional methods rely on gradient-based~\cite{gradient1, gradient2} and statistical approaches~\cite{colorstatistic1, colorstatistic2}.

Recent advances are dominated by deep learning techniques. 
Most learning-based methods adopt U-Net-style encoder-decoder frameworks with task-specific modules~\cite{cong2020dovenetdeepimageharmonization, ling2021regionawareadaptiveinstancenormalization, chen2023hierarchicaldynamicimageharmonization}. 
Several works explore full-resolution harmonization by combining low- and high-resolution branches using learned transformations~\cite{xue2022dccfdeepcomprehensiblecolor, pctnet, meng2024aict}. 
Recently, diffusion models have been introduced for harmonization. Li et al.~\cite{li2023imageharmonizationdiffusionmodel} employ ControlNet with HSV/HSL color transfer, while Chen et al.~\cite{chen2023zero} propose a zero-shot diffusion-VLM pipeline. Zhou et al.~\cite{zhou2024diffharmony, zhou2024diffharmonypp} address latent distortion with a refined stage or Harmony-VAE, achieving state-of-the-art (SOTA) results.

\subsection{Image Harmonization Dataset}

Existing dataset construction methods can be grouped into four categories:
(1) Capture-based: Collecting real images of the same foreground under different backgrounds (e.g., GMSDataset~\cite{GMSDataset}, Hday2night~\cite{cong2020dovenetdeepimageharmonization, day2night});
(2) Manual adjustment: Using professional adjustments of foregrounds to generate variants (e.g., HAdobe5k~\cite{cong2020dovenetdeepimageharmonization}, RealHM~\cite{jiang2021sshselfsupervisedframeworkimage}).
These two methods are often time-consuming and labor-intensive.
(3) Rendering-based: Placing 3D models in virtual scenes to simulate realistic lighting and shadows (e.g., RdHarmony~\cite{cao2022deepimageharmonizationbridging}, IllumHarmony~\cite{hu2024sidnet}), though domain gaps remain a challenge;
(4) Synthetic foreground generation: Creating disharmonious composites by manipulating foreground appearance, with color transfer being a representative approach (e.g., HFlickr and HCOCO in iHarmony4~\cite{cong2020dovenetdeepimageharmonization}).
Our proposed method, Random Poisson Blending falls into the category (4), but introduces richer and more realistic variations in lighting and color, which help improve model generalization to real-harmonization images.

\section{Method}

\subsection{Overall Model Architecture and Data Flow}

Our model is based on the LDM, which consists of two parts: Variational AutoEncoder (VAE) and diffusion network (i.e. U-Net). The VAE includes an Encoder $\mathcal{E}$ and a Decoder $\mathcal{D}$. The Encoder maps an image $I$ to the latent space, $z=\mathcal{E}(I)$, while the Decoder reconstructs the image from the latent space, $\hat{I} = \mathcal{D}(z)$. 
In the first stage, the VAE is trained independently. Then, VAE is locked and the diffusion U-Net is trained in the latent space. 

During the inference phase, a pure noise $z_T$ is first sampled from a Gaussian distribution. Then, the diffusion U-Net iteratively predicts the noise at each time step $t$ and denoises to obtain $z'_0$. Finally, the VAE decoder $\mathcal{D}$ projects back to the pixel space to reconstruct the image.

Given the foreground image $I_f$, the background image $I_b$, and the foreground mask $M$, the composite image $I_c$ can be defined as: $I_c = M \times I_f + (1-M) \times I_b$. 
In our approach, $I_c$ is encoded into latent space, then the harmonization process is carried out in the latent space. Finally, the denoising result is decoded back into pixel space as the image harmonization output.

Compared to LDM, our model has two improvements. 
We replace the original VAE with Clear-VAE, incorporate the Harmony Controller with MACA, and perform fine-tuning.
The following will provide a detailed explanation.

\subsection{Clear-VAE}
\label{sec:clear-VAE}

During VAE encoding process, detail information is lost, leading to a decline in the quality of generated images with unclear details. Therefore, we need to restore these details from the composite images. 
However, directly incorporating content from the original foreground~\cite{yu2024promptfixpromptfixphoto,zhou2024diffharmony,zhou2024diffharmonypp}, while preserving details, also introduces disharmonious elements that negatively impact the generated images.

We propose a new VAE architecture, named {\bf Clear-VAE}.
Preserving high-frequency information in the foreground region helps maintain content details while remaining unaffected by disharmonious factors.
We use \textbf{Adaptive Filter} (AF) to extract high-frequency details from skip-connect features in each layer, and then apply zero-initialized $1\times1$ convolutions to merge them with features in the decoder.
As shown in Fig.~\ref{fig: overview}, AF offers a wide range of filters with adaptive learnable parameters $\alpha$.
Each filter in the AF has its own advantages and disadvantages. Thus, we use the learnable parameters to adaptively combine them together, extracting more types of useful information.
Furthermore, the AF can be reparameterized as one $3\times3$ and one high-pass filter after training to achieve lightweight.
Clear-VAE is trained independently with the contrastive regularization loss (Eq.~\eqref{eq:cr loss}).

\subsection{Harmony Controller with MACA}
\label{sec: harmony controller}

To inject additional image conditions (such as masks, composite images, Canny edges, etc.) in LDM, we add a ControlNet-style~\cite{zhang2023addingconditionalcontroltexttoimage} adaptive encoder as the Harmony Controller (Fig.~\ref{fig: overview}). As a result, more image information can be adaptively injected into the output, while maintaining the stability of the diffusion network.

\begin{figure*}[!t]
    \centering
    \includegraphics[width=1\linewidth]{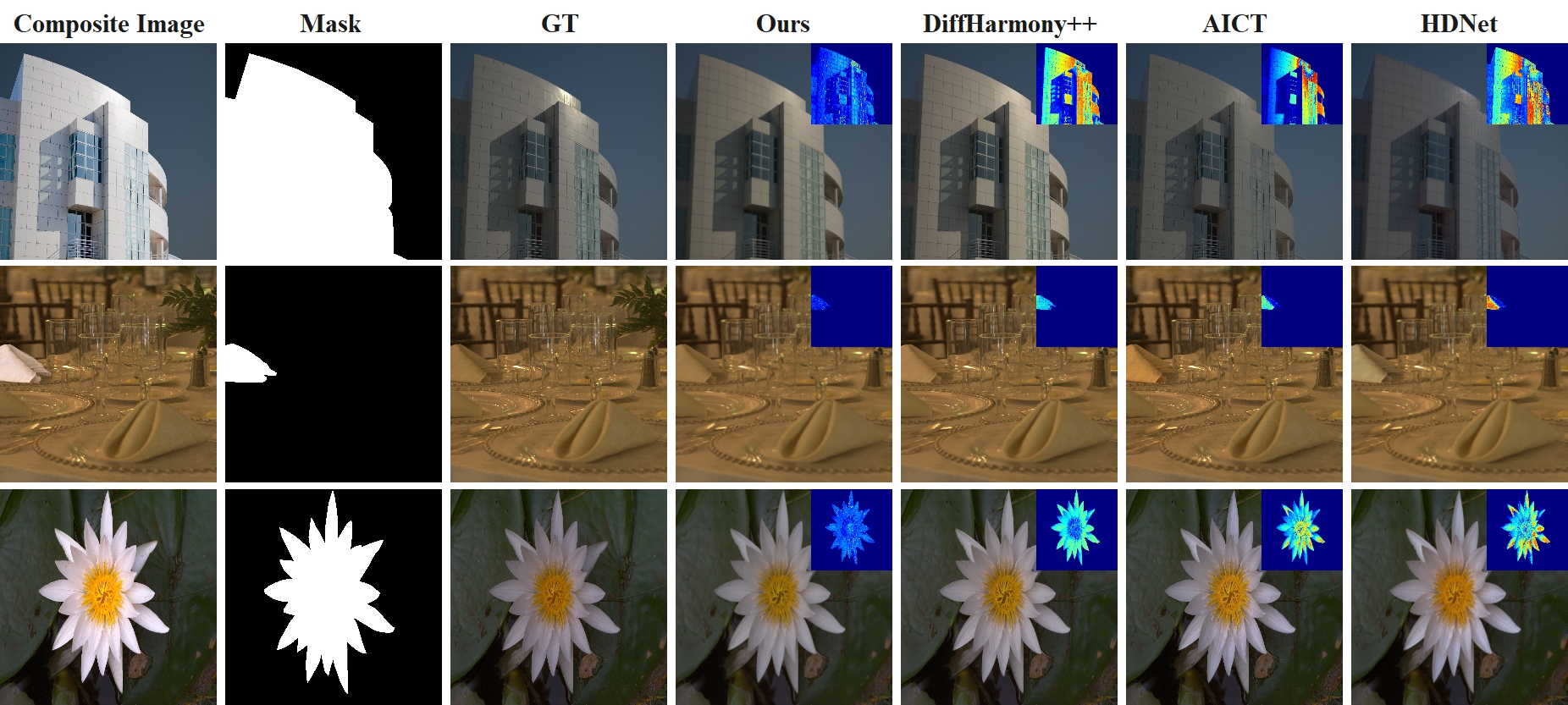}
    \caption{ \textbf{Qualitative comparison results and error maps on the testing dataset of iHarmony4~\cite{cong2020dovenetdeepimageharmonization}.} The error maps are computed based on absolute error.}
    \label{fig:ihar4_images}
\end{figure*}

Self-attention and cross-attention are widely used in LDM to model the importance relationships between pixels and between pixels and text condition~\cite{rombach2022highresolutionimagesynthesislatent}.
However, compared to pixel information, channel information is more suitable for capturing the style of the foreground region. And pixel-level adaptation is unsuitable for low-level color and texture features~\cite{chen2023hierarchicaldynamicimageharmonization}.
Nevertheless, we observe that previous works learn the channel importance either through global pooling applied to the entire feature map~\cite{CBAM,SSAM} or by separately processing the foreground and background regions~\cite{chen2023hierarchicaldynamicimageharmonization}, lacking cross-integration of channel information between them.
Based on these observations, we propose a new module called {\bf M}ask-aware {\bf A}daptive {\bf C}hannel {\bf A}ttention ({\bf MACA}). This module (Fig.~\ref{fig: overview}) adaptively adjusts the importance of channels in the feature map based on the channel information of both the foreground and background.

For the feature map $X\in R^{H\times W \times C}$, we first use a mask $M\in R^{H\times W\times 1}$  (and reverse mask $\overline{M}=1-M$) to obtain the foreground feature map $X_f$ (and the background feature map $X_b$), i.e., $X_f = X\cdot M, \ X_b = X\cdot \overline{M}$.

These two feature maps are then processed through convolution and Global Average Pooling (GAP) to obtain the global channel information $C_f\in R ^{1\times 1\times C}$ and $C_b\in R^{1\times 1 \times C}$, respectively.
The global channel information \( C_f \) and \( C_b \) are concatenated and passed through an MLP, which regresses to obtain the scale factor \( \alpha \in R^{1\times 1\times C}\) and the shift factor \( \beta\in R^{1\times 1\times C} \). The feature map \( X \) is then adjusted using the following formula to obtain \( X_c \):

\begin{equation}
    X_c = X \cdot \alpha + \beta
\end{equation}

Similarly, we use the mask \( M \) to ensure that the adjusted features only inject to the foreground part. Therefore, the output \( Y \) of MACA is represented as:

\begin{equation}
    Y = X_c \cdot M + X_b
\end{equation}

The encoder architecture in LDM (same as the controller) can be divided into four blocks based on the feature map size~\cite{zhang2023addingconditionalcontroltexttoimage}.
At the end of each block, we insert MACA to integrate the channel importance information at that scale and adjust the feature map accordingly. MACA adjusts low-level features such as style, color, and brightness by enhancing or suppressing specific channels.

\subsection{Training and Loss Function}

\label{sec:training}

\noindent \textbf{Training of Clear-VAE} Clear-VAE is trained separately following~\cite{rombach2022highresolutionimagesynthesislatent}. 
We fix the encoder and fine-tune the AF and decoder. 
The supervised training objective is to reconstruct real images.
During training, the encoder encodes both the ground truth $I$ and composite image $I_c$, resulting in $z$ and $z_c$. The skip-connect features $s_c$ from encoding composite images, after high-frequency information extraction via AF, is added to decoding $z$, obtaining the reconstructed images $\hat{I} = \mathcal{D}(z,\text{AF}(s))$. 
In the inference stage, the skip-connect features extracted from the composite image is used to guide the reconstruction of the output of DM.

To measure the difference between the reconstructed image and the ground truth (GT), we use the MSE Loss as the base loss function.

\begin{equation}
    \mathcal{L}_{rec} = \|~I - \hat{I}~\|_2^2
\end{equation}

Using skip-connect features from composite images during the decoding phase can help preserve fine details. However, it also carries the risk of introducing disharmonious information, which may negatively impact the generated foreground.
To address this, we also introduce \textbf{contrastive regularization loss} $\mathcal{L}_{cr}$, aiming to ensure that the reconstructed image is not only close to the positive samples (real images) but also pushed away from the negative samples (composite images).
The formula is similar to the first term of ~\cite{hang2022scscoselfconsistentstylecontrastive}, but we use Random Poisson Blending to generate more composite images as negative data.

\begin{equation}
\label{eq:cr loss}
    \mathcal{L}_{cr} = \frac{D(f,f^+)}{D(f,f^+)+\sum^{K}_{k=1}D(f,f^-)}
\end{equation}

In Eq.~\eqref{eq:cr loss}, $f$,$~f^+$,$~f^-$ are the feature vectors of the foreground region of the reconstructed image, the GT, and the composite images, respectively. The feature vector is extracted from a pre-trained VGG16 network~\cite{vgg}. $D(\cdot)$ is the $\ell_1$ distance function. 
K is set to 3 as the number of negative samples.
The final loss is given by the following equation. To balance the influence of different losses, we empirically set the value of $\lambda$ to 0.3.

\begin{equation}
    \mathcal{L}_{VAE} = \mathcal{L}_{rec} + \lambda \mathcal{L}_{cr}
\end{equation}

\noindent \textbf{Training of LDM} 
For the learning process of harmonization in latent space, where $z_0,c,\epsilon\sim\mathcal{N}(0,1)$, the loss is defined as:

\begin{equation}
\label{eq:latent mask-weight mse}
\mathcal{L}_{LDM}= \frac{M\cdot\|\epsilon - \epsilon_{\theta}(z_t,t,c) \|_2^2}{\max\{A_{min},\sum_{h,w}M_{h,w}\}}
\end{equation}

Eq.~\eqref{eq:latent mask-weight mse} can be seen as the implementation of the foreground MSE loss ~\cite{sofiiuk2020foregroundawaresemanticrepresentationsimage} in the latent space.
$H$, $W$ represent the height and width of the image in the latent space and $M$ is the mask resized to $(H, W)$. 
$A_{min}$ is a hyper-parameter. We set $A_{min} = HW/5$ in all our experiments.

\begin{figure}
    \centering
    \includegraphics[width=0.85\linewidth]{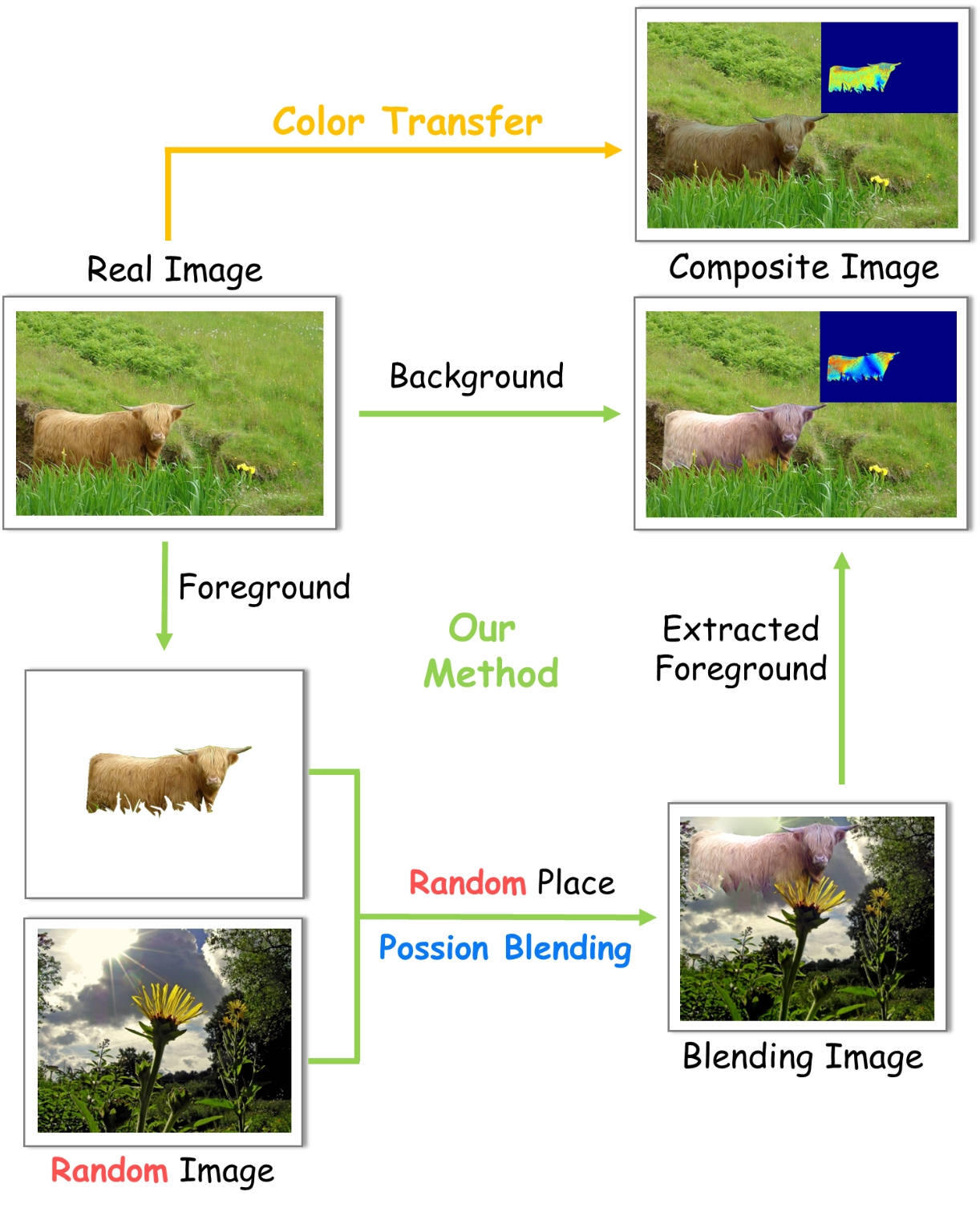}
    \caption{\textbf{Random Poisson Blending Process.} The comparison of composite images shows that our method introduces more diverse local variations.}
    \label{fig:random possion blending}
\end{figure}

Unlike general visual generation, when applying LDM to image harmonization, the generated content is strictly constrained, requiring adjustments to the original style. We found that pretraining the diffusion network specifically for the image harmonization task is necessary.

Therefore, in the first stage, we independently train the U-Net to obtain a diffusion network with image harmonization capabilities. 
In the second stage, we integrate the Harmony Controller (initialized with the weights of the UNet Encoder), lock the U-Net backbone, and fine-tune it. This stage of the training is similar to ControlNet~\cite{zhang2023addingconditionalcontroltexttoimage}.

\section{Dataset Construct}

\subsection{Random Possion Blending}

Color-to-color transformation methods, represented by color transfer, have been widely adopted for synthesized datasets construction~\cite{tsai2017deepimageharmonization,cong2020dovenetdeepimageharmonization}. 
The foregrounds produced by these methods exhibit noticeable global changes but lack local dynamics, which to some extent limits the effectiveness of image harmonization~\cite{ren2024relightfulharmonizationlightingawareportrait}.

To address the aforementioned limitations, we propose a Region-to-Region transformation method, \textbf{Random Poisson Blending}. 
Specifically, we use Poisson Blending~\cite{gradient1} to transfer color and light condition from a random region in a random image to the foreground of a real image, creating more diverse and challenging composite images.

\begin{table*}[!t]
  \centering
  \resizebox{\linewidth}{!}{
    \begin{tabular}{lccccccccccccccc}
    \toprule[1.25pt]
    \multirow{2}[3]{*}{Method} & \multicolumn{3}{c}{HCOCO} & \multicolumn{3}{c}{HAdobe5k} & \multicolumn{3}{c}{HFlickr} & \multicolumn{3}{c}{Hday2night} & \multicolumn{3}{c}{ALL} \\
\cmidrule(r){2-4} \cmidrule(r){5-7} \cmidrule(r){8-10} \cmidrule(r){11-13} \cmidrule(r){14-16}
& PSNR↑ & MSE↓  & fMSE↓ & PSNR↑ & MSE↓  & fMSE↓ & PSNR↑ & MSE↓  & fMSE↓ & PSNR↑ & MSE↓  & fMSE↓ & PSNR↑ & MSE↓  & fMSE↓ \\
\midrule
    Composite & 33.94 & 69.37 & 996.59 & 28.16 & 345.54 & 2051.61 & 28.32 & 264.35 & 1574.37 & 34.01 & 109.65 & 1409.98 & 31.63 & 172.47 & 1376.42 \\
   S$^2$AM & 35.47 & 41.07 & 542.06 & 33.77 & 63.40  & 404.62 & 30.03 & 143.45 & 785.65 & 35.69 & 50.87 & 835.06 & 34.35 & 59.67 & 594.67 \\
    DoveNet  & 35.83 & 36.72 & 551.01 & 34.34 & 52.32 & 380.39 & 30.21 & 133.14 & 827.03 & 35.27 & 51.95 & 1075.71 & 34.76 & 52.33 & 532.62 \\
    RAINNet & 37.08 & 29.52 & 501.17 & 36.22 & 43.35 & 317.35 & 31.64 & 110.59 & 688.40 & 34.83 & 57.40  & 916.48 & 36.12 & 40.29 & 469.60 \\
    iDIH-HRNet  & 39.64 & 14.01 & N/A & 37.35 & 21.36 & N/A & 34.03 & 60.41 & N/A & 37.68 & 50.61 & N/A & 38.31 & 22.00 & 252.00 \\
    DCCF& 39.52 & 14.87 & 272.10 & 37.18 & 23.43 & 172.49 & 33.84 & 61.42 & 411.56 & 38.08  & 45.09 & 655.46 & 38.50 & 22.05  & 265.52 \\
    PCT-Net & 40.78 & 10.72 & 208.26 & 39.97 & 21.25 & 157.24 & 35.13 & 44.30 & 341.10 & 37.65 & 44.74 & 654.81 & 39.85 & 18.16 & 216.25 \\ 
    GKNet & 40.32 & 12.95 & 222.31 & 39.97 & 17.94 & 138.22 & 34.45 & 57.58 & 372.90 & 38.47 & 42.76 & 546.06 & 39.53 & 19.90 & 220.44 \\ 
    HDNet & 41.04 & 11.60 & N/A & 41.17 & {\bf 13.58} & N/A & 35.81 & 47.39 & N/A & 38.85 & 31.97 & N/A & 40.46 & 16.55 & 179.49 \\ 
    AICT & 40.68 & 10.74 & 206.24 & 40.55 & 16.50 & 133.15 & 35.33 & 42.58 & 320.45 & 37.93 & 41.27 & 594.92 & 39.99 & 16.53 & 204.67 \\ 
    \midrule
    DiffHarmony & 41.71 & 9.18 & 170.44 & 41.08 & 19.51 & 120.78 & 37.10 & 30.89 & 216.27 & \underline{39.45} & {\bf 22.42} & \underline{470.846} & 40.97 & 14.86 & 166.48 \\ 
    DiffHarmony++ & \underline{42.42} & \underline{8.43} & \underline{155.73} & \underline{41.78} & 18.81 & \underline{113.18} & \underline{37.74} & \underline{28.77} & \underline{201.52} & \textbf{39.49} & \underline{22.48} & {\bf 464.35} & \underline{41.66} & \underline{13.98} & \underline{153.98} \\ 
    Ours & {\bf 42.62} & {\bf 7.93} & {\bf 146.81} & {\bf 42.22} & \underline{15.70} & {\bf 102.14} & {\bf 38.04} & {\bf 26.28} & {\bf 185.38} & 39.42 & 22.64 &  497.185 & {\bf 41.94} & {\bf 12.51} & {\bf 144.38} \\
    \bottomrule[1.25pt]
    \end{tabular}%
    }

      \caption{\textbf{Quantitative comparisons across four sub-datasets of iHarmony4~\cite{cong2020dovenetdeepimageharmonization}.} $\uparrow$ indicates the higher the better, and $\downarrow$ indicates the lower the better. Best results are in bold and the suboptimal results are in underline. }
  \label{tab:ihar4_metrics}%
  \vspace{-1.0em}
\end{table*}

As shown in Fig.~\ref{fig:random possion blending}, the process of our method is as follows:
(1) Given an image $I_t$ and the foreground mask $M$, we obtain the foreground object $F_t$ from $I_t$, where $F_t$ is the foreground that needs to be adjusted. 
(2) We randomly select a different image $I_r$ from the dataset as the reference image. Then, we randomly choose a region in $I_r$ and apply Poisson blending to blend $F_t$ onto $I_r$, resulting in $I_p$.
(3) Based on the blending position, we extract the new foreground $F_p$ from $I_p$. To control the strength of the Poisson blending, we introduce a hyper-parameter $\alpha$, so that $\tilde{F}_p = \alpha ~\times~ F_p + (1-\alpha) ~\times~ F_t$.
(4) We paste $\tilde{F}_p$ back into its original position to obtain the synthetic image $\tilde{I}_t$.
To ensure the validity of Poisson Blending and the quality of the synthetic image, we perform some additional operations, the details of which are provided in the Supplementary.

Unlike color transfer based on global color statistics, our approach leverages richer low-level background cues—including color patterns, texture, layout, and lighting—that jointly contribute to disharmony in a semantically agnostic manner. 

From the comparison in Fig.~\ref{fig:random possion blending}, it is evident that our method generates composite images with richer local changes, which pose greater challenges for image harmonization. More results generated by our method can be found in the Supplementary Material.

\subsection{Composite Image Generation and Filtering}

We choose the DUTS~\cite{DUTS} and ADE20K~\cite{ade20k} datasets as the base datasets. Both of these datasets were originally created for image segmentation tasks and have high-quality masks. They contain 15,572 and 27,574 image pairs, respectively, covering most daily life scenarios.
For each real image $I_t$, our method generates $N$ alternative synthesized versions as candidates $\{\tilde{I}^{~(i)}_{t}\}_{i=1}^{N}$. We then apply an aesthetic scoring model~\cite{shunk031_aesthetics_predictor_2022} and manual filtering to remove suboptimal synthesized images.

Since our composite image generation is based on \textbf{R}andom \textbf{P}oisson Blending, the dataset is named \textbf{RPHarmony}, which consists of 12,787 training images and 1,422 test images.
Based on the image sources, RPHarmony can be divided into two sub-datasets, namely R-DUTS and R-ADE.
Further details on the synthetic datasets are provided in the Supplementary.

\section{Experiments}

\subsection{Datasets and Metrics}

We conducted experiments on the iHarmony4 dataset~\cite{cong2020dovenetdeepimageharmonization}, comprising four subsets: HCOCO, HAdobe5k, HFlickr, and Hday2night, with 65,742 image pairs for training and 7,404 for testing. 
Each image pair consists of a synthetic composite image, its foreground mask, and a corresponding real image. Following ~\cite{zhou2024diffharmony,zhou2024diffharmonypp}, training images were resized to 512×512. Testing was performed on images at a size of 1024×1024, with results subsequently resized to 256×256 for evaluation.

For the experiments on RPHarmony, we first initialized the models with weights pre-trained on iHarmony4 and then finetune the model on RPHarmony.
Training and testing configurations were identical to those used for iHarmony4. Performance was evaluated using PSNR, MSE, fMSE, and SSIM. 

\subsection{Implementation Details}

The following describes the training configurations for our model on the iHarmony4 dataset. Training details for models involved in the RPHarmony dataset can be found in the Supplementary.

We trained the Clear-VAE model for 30 epochs using the AdamW optimizer with $\beta_1=0.9$ and $\beta_2=0.999$. 
The learning rate of 1e-4 and the batch size of 4 were used, with gradients accumulated over 8 batches (accumulate\_grad\_batches = 8).
LDM training consists of two stages. 
In the first stage, we pretrained the U-Net following the experimental setup of DiffHarmony~\cite{zhou2024diffharmony}.
Secondly, the Controller was trained for 5 epochs on the iHarmony4 training set using the AdamW optimizer, with a learning rate of 1e-5.
Data augmentation included random resized crops and random horizontal flips. 
We set the batch size to 4 with gradient accumulation over 8 batches.
Inference employed the Euler ancestral discrete scheduler~\cite{Euler} with 10 sampling steps. Our models were implemented in PyTorch and trained on three NVIDIA 4090 GPUs.

\subsection{Comparison with Other Methods}

We compared our approach against LDM-based models ~\cite{zhou2024diffharmony,zhou2024diffharmonypp} and other SOTA methods~\cite{SSAM,cong2020dovenetdeepimageharmonization,ling2021regionawareadaptiveinstancenormalization,sofiiuk2020foregroundawaresemanticrepresentationsimage,xue2022dccfdeepcomprehensiblecolor,pctnet,GKNet,chen2023hierarchicaldynamicimageharmonization,meng2024aict}.

\noindent \textbf{Performance comparisons on the iHarmony4.} Fig.~\ref{tab:ihar4_metrics} show our method outperforms existing approaches on almost all metrics across each subset of the test set. Results for other methods were taken from their respective publications or generated using their official code. While our method achieves slightly lower performance on the Hday2night subset, this can likely be attributed to its smaller size and differing image content compared to the other subsets~\cite{pctnet}. Qualitative results are shown in Fig.~\ref{fig:ihar4_images}.

\begin{table}[t]
\centering
\footnotesize
\begin{tabular}{ccccc}
    \toprule[1.25pt]
    Model&PSNR$\uparrow$&MSE$\downarrow$&fMSE$\downarrow$&SSIM$\uparrow$\\\midrule
    Composite & 25.91 & 366.32 & 2362.08 & 0.9580 \\
    PCT-Net& 33.26 & 60.39 & 332.61 & 0.9796 \\
    AICT&33.28 & 60.38& 333.15 & 0.9547\\
    HDNet&34.46&47.52&252.54&0.981\\
    DiffHarmony++&\underline{ 36.03}&\underline{42.16}&\underline{203.45}&\underline{0.9861}\\
    Ours&{\bf 36.32}&{\bf 40.25}&{\bf 192.66}&{\bf0.9872}\\
    \bottomrule[1.25pt]
\end{tabular}

\caption{\textbf{Quantitative comparisons on RPHarmony.} Best results are in bold and the suboptimal results are in underline. }
\label{tab:rph_metrics}
\vspace{-10pt}
\end{table}

\noindent \textbf{Performance comparions on the RPHarmony.} After fine-tuning on the RPHarmony dataset, we compared our method with recent SOTA models~\cite{pctnet,meng2024aict,chen2023hierarchicaldynamicimageharmonization,zhou2024diffharmonypp} (Fig.~\ref{tab:rph_metrics}). The increased foreground variation in RPHarmony presents a greater challenge. Interestingly, several models performing comparably on iHarmony4 showed a significant performance drop on RPHarmony. Our method maintained its superior performance. Qualitative comparisons are shown in the Supplementary.

\subsection{Ablation Study}

\begin{table}[!t]
  \centering
  \footnotesize
  \setlength{\tabcolsep}{4.0pt}
    \begin{tabular}{crcccc}
    \toprule[1.25pt]
    \multicolumn{1}{c}{\multirow{2}[2]{*}{Model}} & \multicolumn{1}{c}{\multirow{2}[2]{*}{Params}} & \multicolumn{2}{c}{Real Images} & \multicolumn{2}{c}{Generated Images} \\
    \cmidrule(r){3-4} \cmidrule(r){5-6}
          &  & \multicolumn{1}{l}{MSE$\downarrow$} & \multicolumn{1}{l}{fMSE$\downarrow$} & \multicolumn{1}{l}{MSE$\downarrow$} & \multicolumn{1}{l}{fMSE$\downarrow$} \\ 
    \midrule
    SD-VAE   & 83.65 M & 22.94      & 678.73      &    13.51 & 160.01     \\
    Harmony-VAE     &  +36.1\text{M}   & 1.72      & 29.66      &  12.63 & 147.03    \\
    \midrule
    w/o AF     &  +1.90\text{M}  & 1.52      & 25.82      &   12.58 & 146.76      \\
    w/o $\mathcal{L}_{cr}$ & +1.90\text{M} & 1.78      &  32.30   & 12.66 & 147.13     \\
    \midrule
    Ours   & +1.90\text{M}  &  {\bf 1.44}     & {\bf 24.68}      & \textbf{12.51} & {\bf 144.38}       \\
    \bottomrule[1.25pt]
    \end{tabular}
      \caption{\textbf{Performance of different VAE design.} Params represents the model's parameter count, where "+xM" indicates an increase of xM parameters compared to the SD-VAE.}
  \label{tab:components_vae}%
  \vspace{-1.0em}
\end{table}%

\begin{table}[t]
    \centering
    \footnotesize
    \begin{tabular}{c|c|ccc}
        \toprule
        \textbf{Datset} & \textbf{Variant} & \textbf{PSNR$\uparrow$} & \textbf{MSE$\downarrow$} & \textbf{fMSE$\downarrow$} \\
        \midrule
        \multirow{3}{*}{\textbf{iHarmony4}} 
        & w/o Controller   & 41.76 & 13.44 & 149.77 \\
        & w/o MACA & 41.92 & 12.55& 145.02 \\
        & Ours                & \textbf{41.94} & \textbf{12.51} & \textbf{144.38} \\
        \midrule
        \multirow{3}{*}{\textbf{ccHarmony}} 
        & w/o Controller  & 41.54& 20.88&  168.54  \\
        & w/o MACA & \textbf{41.66} & 22.07 & 172.24 \\
        & Ours               & 41.57 & \textbf{20.80} & \textbf{165.99} \\
        \bottomrule
    \end{tabular}
     \caption{\textbf{Quantitative ablation study of our model variants on iHarmony4~\cite{cong2020dovenetdeepimageharmonization} and ccHarmony~\cite{ccHarmony}.} 
     The best results are in bold}
    \label{tab:maca-abla}
    \vspace{-10pt}
\end{table}

\begin{table}[t]
  \centering
  \begin{tabular}{c|cccc}
    \toprule[1.25pt]
    Metric  & HDNet &HDNet* & Ours & Ours* \\
    \midrule
    Quality Score $\uparrow$  & 3.824 & 3.877 & 4.039 & \textbf{4.082} \\
    \bottomrule[1.25pt]
  \end{tabular}
\caption{\textbf{Quantitative results on the RealHM~\cite{jiang2021sshselfsupervisedframeworkimage}.} The quality score is computed based on the DeQA-Score model~\cite{deqa_score}. * denotes models that are fine-tuned on the RPHarmony dataset. Best results are in bold.}
  \label{tab:deqa-main}
\end{table}

\begin{figure}[t!]
    \centering
    \captionsetup[subfloat]{labelformat=empty}
    \resizebox{\linewidth}{!}{
    \subfloat[\fontsize{11}{13}\selectfont  Input]{
    \begin{minipage}[b]{0.2\linewidth}
    \includegraphics[width=1\linewidth]{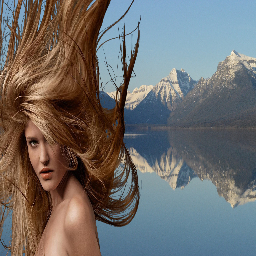}\vspace{1pt}
    \includegraphics[width=1\linewidth]{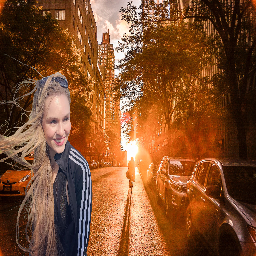}\vspace{1pt}

    \end{minipage}}
\subfloat[\fontsize{11}{13}\selectfont 
 Mask]{
    \begin{minipage}[b]{0.2\linewidth}
    \includegraphics[width=1\linewidth]{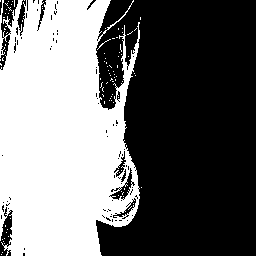}\vspace{1pt}
    \includegraphics[width=1\linewidth]{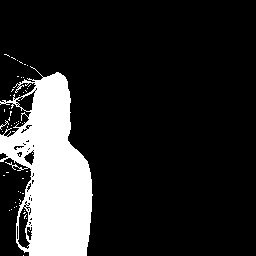}\vspace{1pt}
    \end{minipage}}
\subfloat[\fontsize{11}{13}\selectfont 
 HDNet]{
    \begin{minipage}[b]{0.2\linewidth}
    \includegraphics[width=1\linewidth]{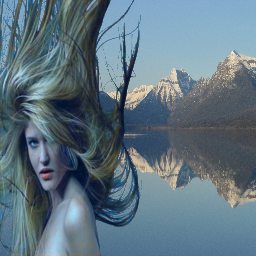}\vspace{1pt}
    \includegraphics[width=1\linewidth]{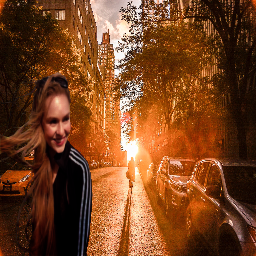}\vspace{1pt}
    \end{minipage}}
\subfloat[\fontsize{11}{13}\selectfont  HDNet*]{
    \begin{minipage}[b]{0.2\linewidth}
    \includegraphics[width=1\linewidth]{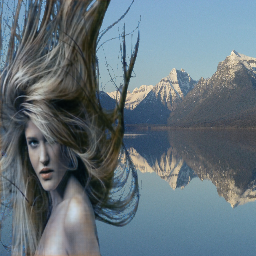}\vspace{1pt}
    \includegraphics[width=1\linewidth]{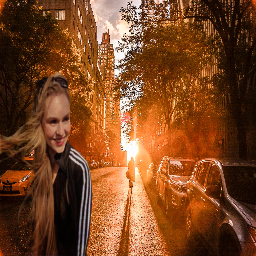}\vspace{1pt}
    \end{minipage}}
\subfloat[\fontsize{11}{13}\selectfont  Ours]{
    \begin{minipage}[b]{0.2\linewidth}
    \includegraphics[width=1\linewidth]{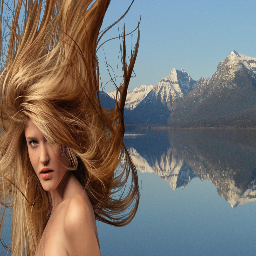}\vspace{1pt}
    \includegraphics[width=1\linewidth]{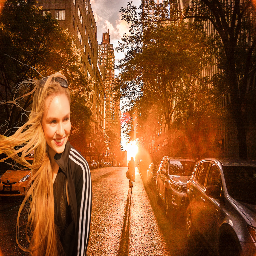}\vspace{1pt}
    \end{minipage}}
\subfloat[\fontsize{11}{13}\selectfont  Ours*]{
    \begin{minipage}[b]{0.2\linewidth}
    \includegraphics[width=1\linewidth]{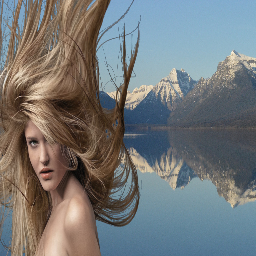}\vspace{1pt}
    \includegraphics[width=1\linewidth]{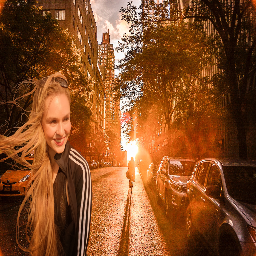}\vspace{1pt}
    \end{minipage}}}
    
    \caption{\textbf{Qualitative results on the RealHM~\cite{jiang2021sshselfsupervisedframeworkimage}} * denotes models that are fine-tuned on the RPHarmony dataset.}
    \label{fig:real-har}
\end{figure}

\noindent \textbf{Evaluation of Clear-VAE.} Fig.~\ref{tab:components_vae} examines the impact of Adaptive Filter (AF) and different loss functions on reconstruction quality. We compared several VAE models, including variants of our proposed method, evaluating both the reconstruction of real images (at the training resolution of 256$\times$256) and the generation of images from diffusion models (with an input resolution of 1024$\times$1024). 
The results clearly demonstrate that high-frequency information extracted by the AF contributes to improved visual consistency. 
Furthermore, the contrastive regulation loss mitigates the influence of disharmonious factors. Removing either component degrades performance.
Notably, Clear-VAE is lightweight, introducing only about 2\% additional parameters compared to SD-VAE.

\noindent \textbf{Evaluation of MACA Module and Harmony Controller.}
Tab.~\ref{tab:maca-abla} shows that the Harmony Controller helps align the generative process with the composite input, ensuring semantic consistency and overall performance. The MACA module further improves results, especially on the ccHarmony dataset, by applying mask-aware attention to better handle disharmony region. Together, they enhance harmonization quality and model generalization.

\noindent \textbf{Effect of Fine-tuning on the RPHarmony Dataset.}
We conduct a comparison between the lightweight model HDNet~\cite{chen2023hierarchicaldynamicimageharmonization} and our model, evaluating their performance on the RealHM dataset~\cite{jiang2021sshselfsupervisedframeworkimage} before and after fine-tuning on RPHarmony. 
Fine-tuned models generalize better to real scenarios, adapting to realistic lighting over simple color alignment, as supported by qualitative results (Fig.~\ref{fig:real-har}) and quality score improvements (Tab.~\ref{tab:deqa-main}).
More results and details are provided in the Supplementary.

\section{Conclusion}

In this work, we introduce the Region-to-Region transformation, which injects information from appropriate regions into the foreground, enabling both better harmonization and the generation of new composite data. 
Building upon this, we propose R2R, a novel model that incorporates Clear-VAE with Adaptive Filter for preserving high-frequency details and eliminating disharmonious elements, as well as the Harmony Controller with Mask-aware Adaptive Channel Attention (MACA) to dynamically adjust the foreground based on both foreground and background channel importance.
To further improve the quality of training data, we propose Random Poisson Blending, a region-to-region transformation method for composite image generation, and construct RPHarmony dataet, which better captures complex real-world lighting conditions. 
Extensive experiments demonstrate the effectiveness of our method. Additionally, models trained on RPHarmony produce more visually harmonious images in real-world scenarios.
Considering the current limitation of model size, future work will explore lightweight diffusion models to improve efficiency. Additionally, we plan to apply Random Poisson Blending to construct larger-scale datasets and investigate the relationship between dataset scale and model generalization.

\bibliography{aaai2026}

\end{document}


\maketitle



In this document, we provide additional materials to support our main paper. 
\textbf{Section 1} details the implementation of synthetic image generation using Random Poisson Blending. 
\textbf{Section 2} provides further descriptions of the base datasets, DUTS and ADE20K, including an analysis of foreground ratios and our filtering process for invalid synthetic images. 
\textbf{Section 3} presents the experimental setup and implementation details for the RPHarmony dataset.
In \textbf{Sections 4 and 5}, we provide additional quantitative and qualitative results respectively to further support the effectiveness of our model.
\textbf{Section 6} highlights the significance of the RPHarmony dataset, supplemented with further quantitative and qualitative evidence demonstrating its effectiveness.
In \textbf{Section 7}, we provide an additional analysis of the computational efficiency of the MACA module.
Finally, in \textbf{Section 8}, we discuss the limitations and future work of our approach.

\section{1\quad Implementation Details and Visual Examples of Random Poisson Blending}
\label{sec:details of RPB}

\subsection{Implementation Details}

In the main paper, we proposed a novel data synthesis method, \textbf{Random Poisson Blending}, for constructing image harmonization datasets. Here, we provide additional implementation details. 
Our method builds upon Poisson blending ~\cite{gradient1}, a classic optimization-based image blending technique originally designed to seamlessly integrate foreground objects into background images. 
This technique aims to preserve foreground content while smoothing transition boundaries. However, a known issue with Poisson blending is that the background image's lighting and color significantly influence the foreground's appearance. Our method leverages this characteristic to synthesize foreground images under diverse background conditions.

Various open-source implementations of Poisson blending exist. For stability and speed, we used pytorch-poisson-image-editing \cite{pytorch_poisson_editing}. Similar to the \textit{NORMAL\_CLONE} mode of OpenCV's \textit{seamlessClone} function, the foreground gradient is always used within the blended region. To improve the quality of synthesized images, we implemented the following optimizations: 
(1) Size and region checks are performed during the random selection of reference images and blending positions to ensure valid Poisson blending operations. 
(2) The blending strength, controlled by the hyperparameter $\alpha$ (as described in [cite the source mentioning $\alpha$]), is randomly sampled between 0.6 and 1.0.

\subsection{Visual Examples and Comparisons}

\begin{figure}[t]
    \centering
    \includegraphics[width=1\linewidth]{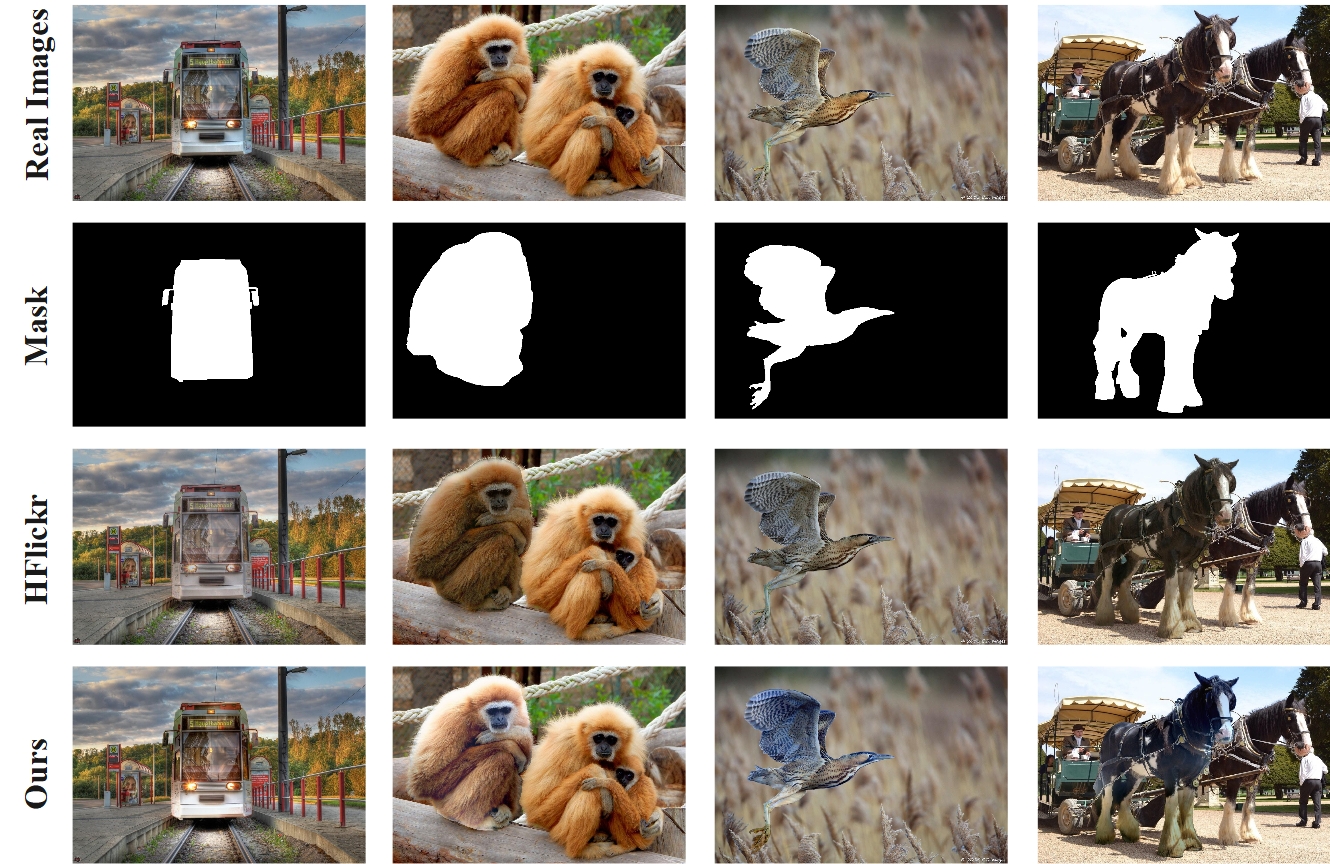}
    \caption{Comparison of synthesized images generated from real HFlickr images using our method and color transfer (i.e., composite images in HFlickr).}
    \label{fig:compass-rpb-ct}
\end{figure}

We apply our method to the real images from the HFlickr dataset, and compare the results with the corresponding composite images generated by color transfer, as illustrated in Fig.~\ref{fig:compass-rpb-ct}. 
It can be observed that our method introduces more diverse local color and illumination variations when generating disharmonious content.

\begin{table*}[t]
  \centering
  {
    \begin{tabular}{ccccccc}
    \toprule[1.25pt]
    \multirow{2}[3]{*}{Method} & \multicolumn{3}{c}{R-DUTS} & \multicolumn{3}{c}{R-ADE20K}  \\
\cmidrule(r){2-4} \cmidrule(r){5-7}
& PSNR↑ & MSE↓  & fMSE↓ & PSNR↑ & MSE↓  & fMSE↓ \\
\midrule
    composite images & 24.82 & 384.91 & 1488.94 & 27.23& 343.86 & 3416.89  \\
    PCT-Net~\cite{pctnet}  & 31.02 & 76.11 & 302.63 & 35.96 & 41.40 & 368.83  \\
    AICT~\cite{meng2024aict}  & 31.04 & 76.30& 302.87 & 35.97 & 41.14 & 369.74  \\
    HDNet~\cite{chen2023hierarchicaldynamicimageharmonization}  & 32.31 & 56.89 & 217.46 & 37.05 & 36.21 & 294.92  \\
    DiffHarmony++~\cite{zhou2024diffharmonypp}  & 33.12 & 55.34 & 201.21 & 39.55 & 26.23 & 206.17  \\
    Ours  & \textbf{33.34} & \textbf{53.00} & \textbf{193.20} & \textbf{39.91} & \textbf{24.85} & \textbf{192.00}  \\

    \bottomrule[1.25pt]
    \end{tabular}%
    }
      \caption{\textbf{Quantitative comparisons on R-DUTS and R-ADE20K.} Best results are in bold.}
  \label{tab:rpharmony_subset}%
  \vspace{-1.0em}
\end{table*}

\section{2\quad Analysis of RPHarmony Dataset}
\label{sec:details of dataset}

\subsection{Composition of RPHarmony Dataset}

\begin{figure}[t]
    \centering
    \includegraphics[width=1.0\linewidth]{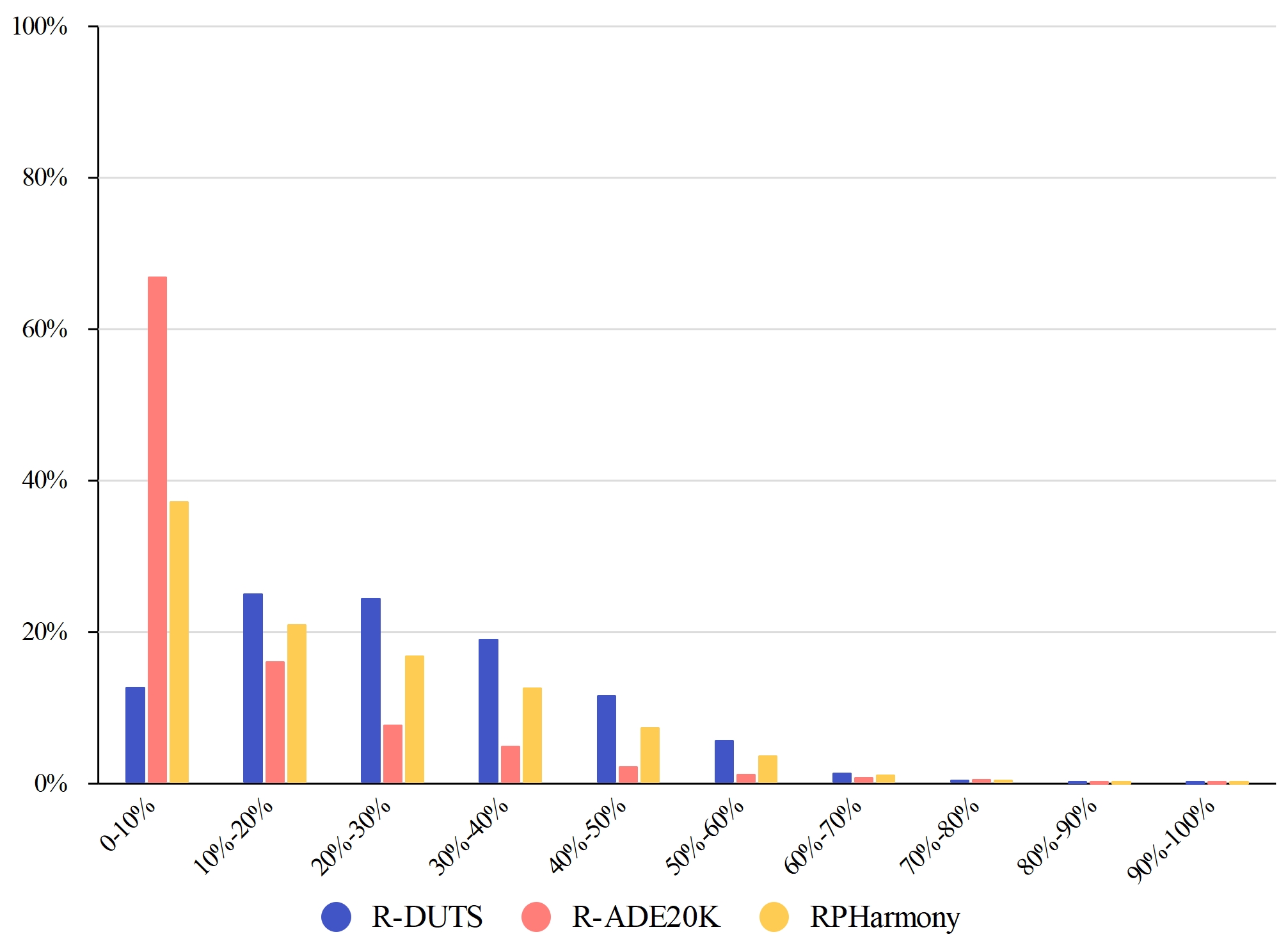}
    \caption{\textbf{The distributions of foreground ratios.} A comparison of the foreground ratio distributions between the RPHarmony dataset and its two sub-datasets, R-DUTS and R-ADE20K. R-DUTS contains a higher proportion of large foreground regions. In contrast, the foreground ratio in R-ADE20K is generally smaller, resembling the distribution seen in the iHarmony4 dataset~\cite{cong2020dovenetdeepimageharmonization}.}
    \label{fig:fg_ratio}
\end{figure}

\begin{figure*}[t]
\centering
\includegraphics[width=1\linewidth]{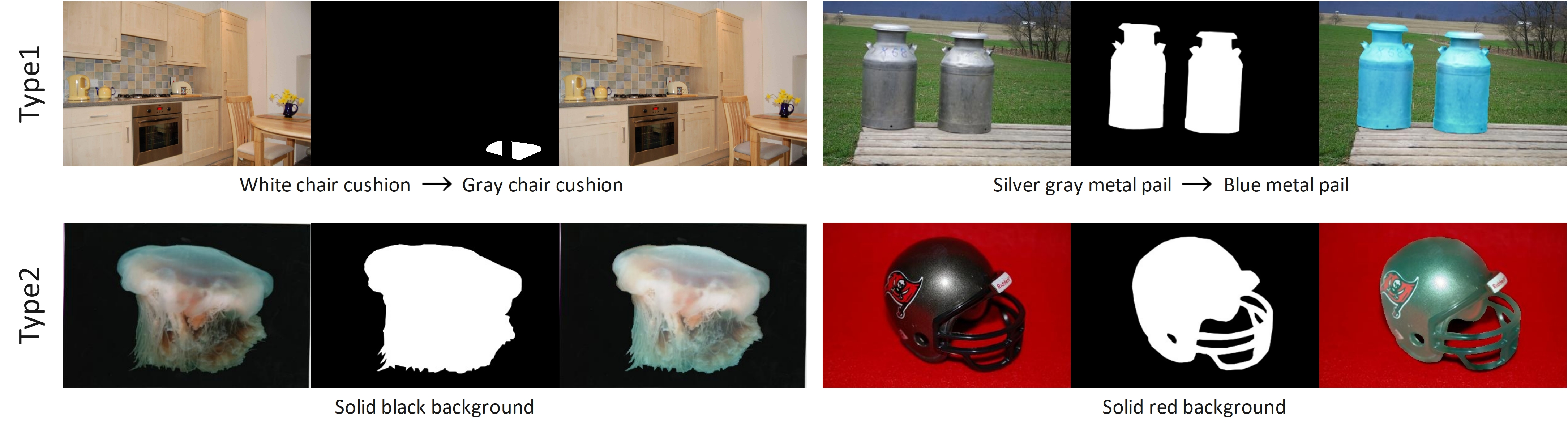}
\caption{\textbf{Examples of manually discarded unqualified composite images.} Two types of unqualified composite images, along with their corresponding real images and masks, will be explained in detail below.}
\label{fig:bad-case}
\end{figure*}

The RPHarmony dataset is derived from two base datasets, DUTS and ADE20K. 
Applying Random Poisson Blending to these datasets yields R-DUTS~\cite{DUTS} and R-ADE20K~\cite{ade20k}, respectively, which together comprise the final RPHarmony dataset. R-DUTS contains 6,999 training images and 778 test images, while R-ADE20K contains 5,788 training images and 644 test images. In total, RPHarmony consists of 12,787 training images and 1,422 test images.

DUTS, originally a saliency detection dataset, comprises 10,553 training images and 5,019 test images. The training images are sourced from the ImageNet DET training/validation sets, while the test images are drawn from the ImageNet DET test set and the SUN dataset. 
ADE20K is a more comprehensive dataset with 27,574 images (25,574 for training, 2,000 for testing) from the SUN and Places databases. It includes annotations for 707,868 unique objects and 193,238 object parts, spanning 365 different scenes. ADE20K is widely used for tasks such as scene parsing, image segmentation, and depth estimation.

The DUTS dataset provides only salient foreground masks, while ADE20K includes masks for multiple objects within each image. To maintain consistency with the foreground ratios observed in the HCOCO ~\cite{cong2020dovenetdeepimageharmonization} dataset, we randomly selected a mask with a foreground ratio between 1\% and 80\% for each image in ADE20K to serve as the foreground mask.

Regarding foreground content, DUTS offers richer and more diverse foregrounds, often featuring living objects, whereas ADE20K primarily contains non-living objects within scenes. While image harmonization focuses on adjusting foreground style rather than deep semantic understanding, we combined these two datasets to generate composite images and ensure greater data diversity.

Regarding foreground ratios (Fig~\ref{fig:fg_ratio}), R-DUTS exhibits a larger and more evenly foreground ratio, closer to that of the HumanHarmony dataset ~\cite{zhou2024diffharmonypp}. Conversely, R-ADE20K primarily features smaller foreground objects, similar to the distribution in iHarmony4 ~\cite{cong2020dovenetdeepimageharmonization}. Overall, the foreground ratio distribution in RPHarmony closely resembles that of Hflickr ~\cite{cong2020dovenetdeepimageharmonization}.

\subsection{Synthetic Images Filtering Process}

For each real image, we generated five candidate synthetic images. These candidates were initially filtered using a neural network model, followed by manual inspection and removal of unsuitable examples.

We used the \cite{shunk031_aesthetics_predictor_2022} (aesthetics-predictor-v2) (a popular aesthetic scoring model available on Hugging Face) for initial candidate evaluation. This CLIP-based model utilizes CLIP image embeddings as input to a final MLP layer for predicting aesthetic scores~\cite{schuhmann2022aesthetic}. 
The first and second highest-scoring composite images for each real image were then manually reviewed.

During manual review, we filtered out unsuitable synthetic images for two primary reasons: (1) Excessive color changes in the foreground, rendering harmonization overly difficult or ambiguous. This often occurred with white or transparent foreground objects. (2) Overly simple or solid-color backgrounds, which lack sufficient contextual information for meaningful foreground harmonization.

Fig.~\ref{fig:bad-case} illustrates examples of excluded synthetic images. In the first row, the left image shows a chair cushion that, while originally white, appears gray due to the blending process, introducing ambiguity. Similarly, the right image shows a metal pail that could be interpreted as either color-shifted due to blue lighting or inherently blue, making harmonization ambiguous. In the second row, the synthetic image features a solid black or red background. Such backgrounds lack realistic contextual information, likely arising from post-processing in the source images to emphasize the foreground. Therefore, we excluded these examples.

\begin{table*}[t]
  \centering
  \setlength{\tabcolsep}{4.0pt}
    \begin{tabular}{cccccccc}
    \toprule[1.25pt]
    \multicolumn{1}{c}{\multirow{2}[2]{*}{Model}} &  \multicolumn{4}{c}{1024$\times$1024} & \multicolumn{2}{c}{4K} \\
    \cmidrule(r){2-5} \cmidrule(r){6-7}
          & \multicolumn{1}{l}{PSNR$\uparrow$} & \multicolumn{1}{l}{MSE$\downarrow$} & \multicolumn{1}{l}{Memory$\downarrow$} & \multicolumn{1}{l}{Inference Time$\downarrow$}& \multicolumn{1}{l}{PSNR$\uparrow$} & \multicolumn{1}{l}{MSE$\downarrow$} \\ 
    \midrule

    PCT-Net   & 39.39      & 22.57      &    910 MB & \textbf{15.2 ms} & 39.83 & 19.36   \\
    AICT    & 39.67    & 19.50      &  \textbf{864 MB} & 17.8 ms  & 40.16 & 16.79   \\
    \midrule
    Ours    &  \textbf{41.49}    & \textbf{16.69}      & 12104 MB & 1269 ms    & \textbf{40.56}& \textbf{16.02}   \\
    \bottomrule[1.25pt]
    \end{tabular}
      \caption{\textbf{Quantitative results on the HAdobe5k dataset~\cite{cong2020dovenetdeepimageharmonization} at high resolutions.} Best results are in bold.}
  \label{tab:high-resolution}%
\end{table*}%

\section{3\quad Implementation Details of the Experiments on the RPHarmony Dataset.}
\label{sec:details of rpharmony exp}

For AICT~\cite{meng2024aict}, PCTNet~\cite{pctnet}, and HDNet~\cite{chen2023hierarchicaldynamicimageharmonization}, we adopted the same training settings used on iHarmony4 and initialized the models with pre-trained weights from iHarmony4. These models were then trained on the RPHarmony dataset for 37, 40, and 38 epochs, respectively.

\begin{figure*}
    \centering
    \includegraphics[width=1\linewidth]{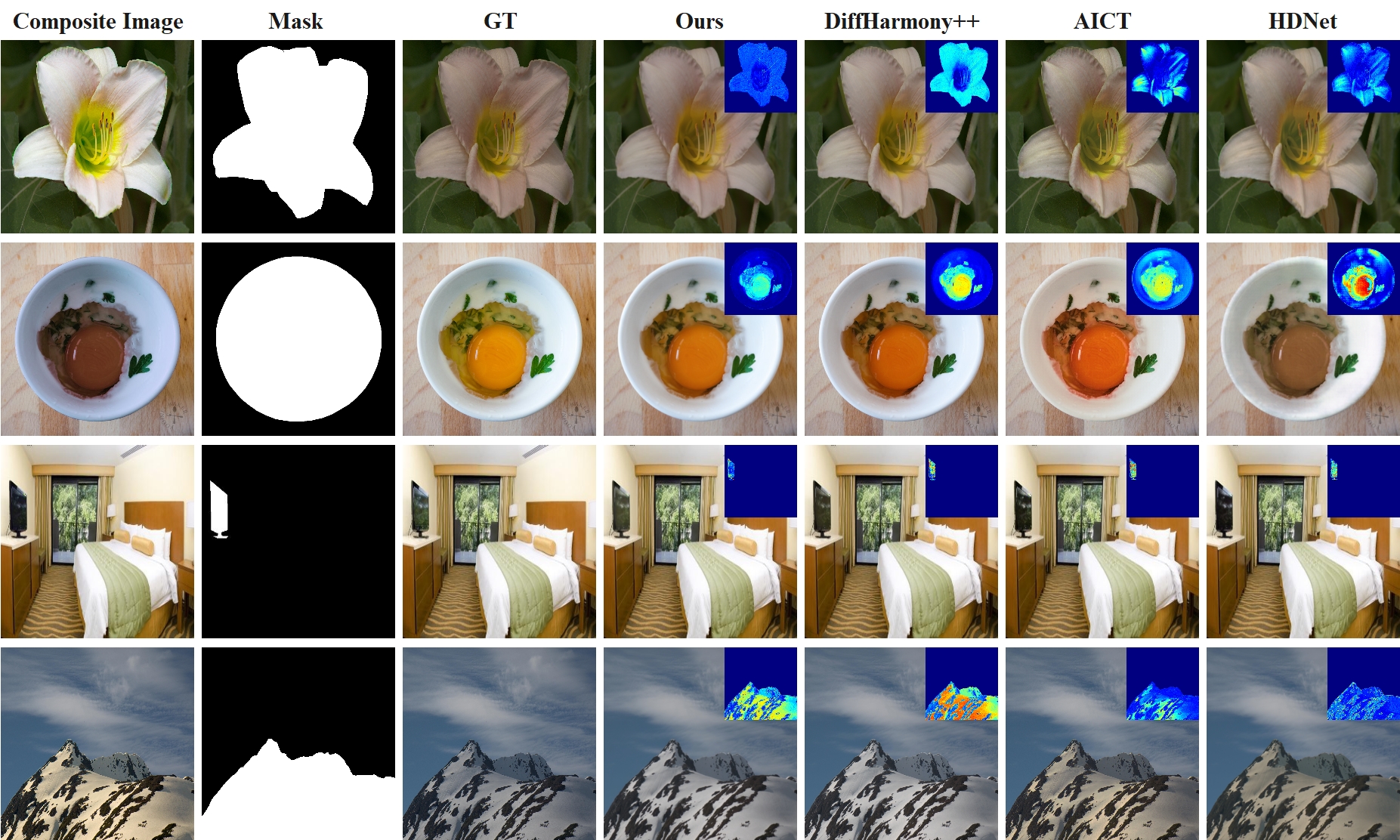}
    \caption{\textbf{Qualitative results on the iHarmony4 dataset~\cite{cong2020dovenetdeepimageharmonization}.} The error maps are computed based on absolute error.}
    \label{fig:ihar4-supp}
\end{figure*}

\begin{figure*}
    \centering
    \includegraphics[width=1\linewidth]{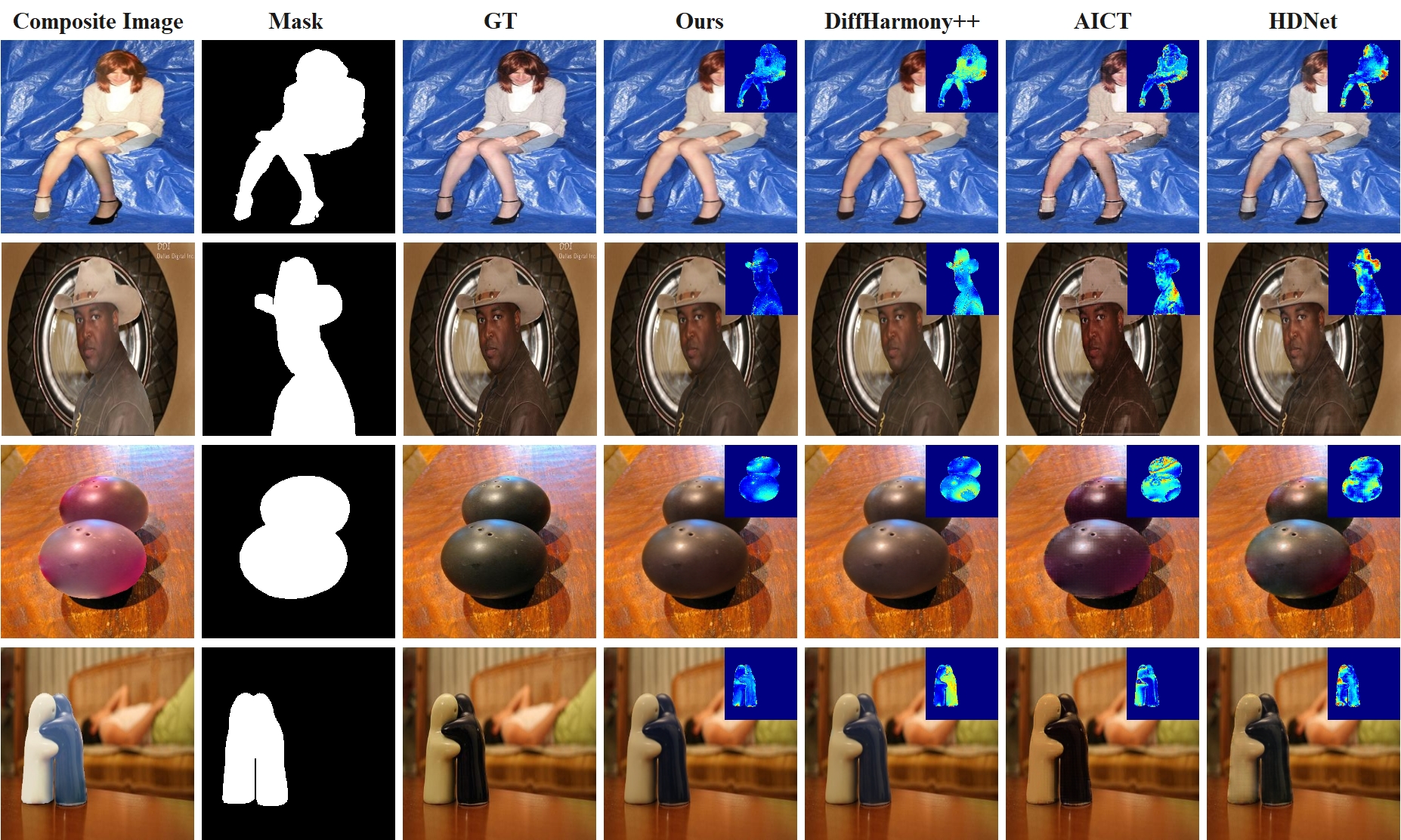}
    \caption{\textbf{Qualitative results on the RPHarmony dataset.} The error maps are computed based on absolute error.}
    \label{fig:rph-supp}
\end{figure*}

We first fine-tuned the U-Net for 40 epochs. This fine-tuned U-Net, combined with the (unmodified) HarmonyVAE, constitutes the fine-tuned DiffHarmony++. 
For our model, the fine-tuned U-Net serves as the initialization for the first training stage. We then add the Harmony Controller and train for a further 5 epochs in the second stage. 
Clear-VAE, like HarmonyVAE, is not fine-tuned.
Both DiffHarmony++ and our model retain the original iHarmony4 experimental settings.

\section{4\quad Supplementary Quantitative Results}

In this section, we provide supplementary quantitative results to further demonstrate the advantages of our proposed method.

\subsection{Performance on the High-resolution Images}

Following the experimental settings described in Appendix A.1 of AICT~\cite{meng2024aict}, we evaluate the performance of our method against state-of-the-art full-resolution image harmonization models (PCT-Net~\cite{pctnet} and AICT~\cite{meng2024aict}) on the HAdobe5k dataset. First, we resize all images in the dataset to 1024×1024 for inference and testing. Next, we select a subset of images with resolutions exceeding 4K (4096$\times$2160) for further evaluation. To address memory constraints, we resize all images to 2048×2048 during inference with our method, and then compute the evaluation metrics at their original resolution. The quantitative metrics, including performance scores, inference time, and memory usage, are summarized in Tab.~\ref{tab:high-resolution}. 

Although our model is designed for general image harmonization rather than specifically for high-resolution scenarios, it still achieves state-of-the-art performance in terms of image quality metrics. Despite being based on a large LDM architecture, our method maintains acceptable memory usage and inference time.

\subsection{Results on the ccHarmony}

Following previous work, we fine-tune the model pre-trained on the iHarmony4 dataset on the ccHarmony dataset~\cite{ccHarmony} before evaluation. 
Specifically, we fine-tune the U-Net for 15 epochs, and then use its weights to initialize the Harmony Controller, which is further fine-tuned for another 10 epochs. 
We compare our method with DiffHarmony++~\cite{zhou2024diffharmonypp}, DCCF~\cite{xue2022dccfdeepcomprehensiblecolor}, GiftNet~\cite{ccHarmony}, AICT~\cite{meng2024aict}. The quantitative results, as shown in Tab.~\ref{tab:ccharmony}.

\begin{table}[t]
  \centering
  \begin{tabular}{cccc}
    \toprule[1.25pt]
    Metric & PSNR$\uparrow$ & MSE$\downarrow$ & fMSE$\downarrow$ \\
    \midrule
    DCCF & 36.62 & 29.25 & 259.83 \\
    GiftNet & 37.59 & 24.55 & 235.20  \\
    AICT & 38.46 & 24.14 & 232.66 \\
    DiffHarmony++ & 41.41 & 22.44 & 171.42 \\
    Ours & \textbf{41.57} & \textbf{20.80} & \textbf{165.99} \\
    \bottomrule[1.25pt]
  \end{tabular}
\caption{\textbf{Quantitative results on the ccHarmony~\cite{ccHarmony}.}  Best results are in bold.}
  \label{tab:ccharmony}
\end{table}

\begin{table*}[th]
  \centering
  \begin{tabular}{c|cccccc}
    \toprule[1.25pt]
    Metric & Composite & MA & HDNet &HDNet* & Ours & Ours* \\
    \midrule
    Quality Score $\uparrow$ & 3.957 & 3.855 & 3.824 & 3.877 & 4.039 & \textbf{4.082} \\
    \bottomrule[1.25pt]
  \end{tabular}
\caption{\textbf{Quantitative results on the RealHM~\cite{jiang2021sshselfsupervisedframeworkimage}.} * denotes models that are fine-tuned on the RPHarmony dataset. ``MA'' stands for manual adjustment results, referring to the harmonized images in the RealHM dataset created via manual editing in Photoshop. Best results are in bold. The quality score is computed based on the DeQA-Score model~\cite{deqa_score}.}
  \label{tab:deqa}
\end{table*}

\section{5\quad Supplementary Qualitative Results}

In this section, we provide supplementary Qualitative results to support our method.
Additional qualitative results along with error maps on the iHarmony4 dataset are presented in Fig.~\ref{fig:ihar4-supp}.
Furthermore, qualitative comparisons between our method and recent state-of-the-art approaches on the RPHarmony dataset are illustrated in Fig.~\ref{fig:rph-supp}.

\begin{figure*}
    \centering
    \includegraphics[width=1\linewidth]{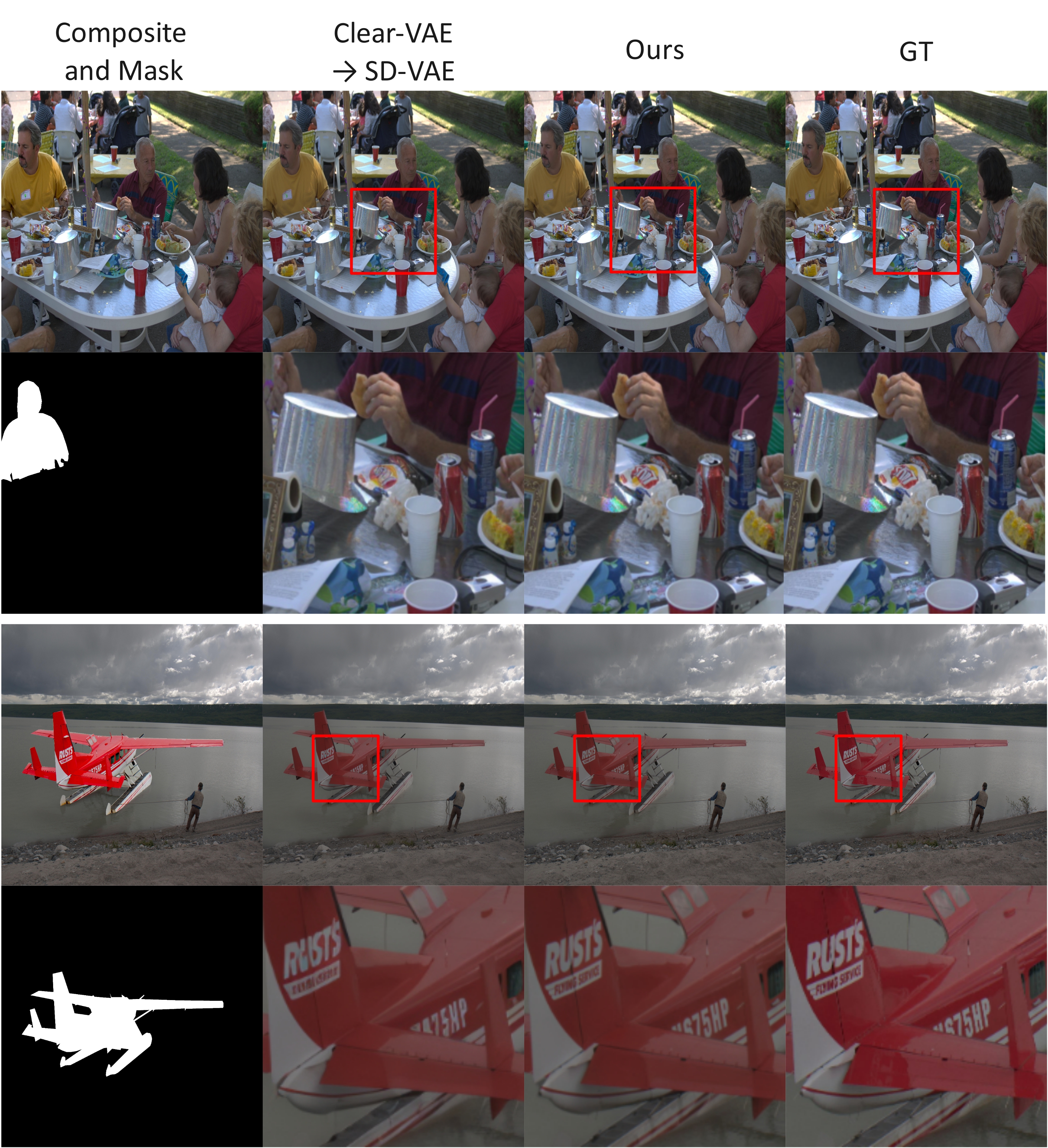}
    \caption{\textbf{High-Resolution Image Harmonization Results.} Zoomed-in regions from the red boxes are shown in the images below for better visualization. To better showcase the model's reconstruction, the backgrounds are not pasted from the original inputs, as is usually done in image harmonization task.}
    \label{fig:high_res}
\end{figure*}

We also provide additional results on high-resolution images, as shown in Fig.~\ref{fig:high_res}. Benefiting from the high-frequency detail preservation capability of Clear-VAE, artifacts commonly seen in high-resolution harmonization are effectively reduced, and fine details such as textures and text are accurately reconstructed. 
Notably, Clear-VAE is lightweight, introducing only about 2\% additional parameters compared to SD-VAE~\cite{rombach2022highresolutionimagesynthesislatent}.

\section{6\quad Effectiveness of RPHarmony: Evaluation on Real Composite Images}

To highlight the impact and general applicability of the RPHarmony dataset, we conducted experiments on real composite images from the RealHM dataset~\cite{jiang2021sshselfsupervisedframeworkimage} using two models of significantly different scales: the lightweight HDNet (~10M parameters) and our larger model (~1.5B parameters), evaluating their performance before and after fine-tuning on RPHarmony.

For the \textbf{quantitative evaluation}, we adopt DeQA-Score~\cite{deqa_score}, a state-of-the-art MLLM-based image quality assessment (IQA) method, to compare model performance on RealHM. As shown in Tab.~\ref{tab:deqa}, fine-tuning on RPHarmony consistently improves model performance on real composite images, indicating enhanced generalization to complex, real-world scenarios.

\textbf{Qualitative results} on RealHM are presented in Fig.~\ref{fig:real-hr2}. After fine-tuning, both models produce more visually harmonious and realistic results, demonstrating better adaptation to real-world backgrounds and lighting conditions.

Building on the observed improvements in both quantitative and qualitative results, we further draw the following insights:
\begin{itemize}
    \item By introducing more complex variations in lighting and color, RPHarmony enables the model to better disentangle appearance from content. As a result, the model adapts the foreground based on the visual context of the background—such as color and illumination—while preserving semantic consistency, rather than merely matching color or style.
    
    \item HDNet, a lightweight yet competitive model trained solely on synthetic harmonization datasets (i.e., iHarmony4), tends to overfit and perform poorly on real composite images, with quality scores in some cases lower than the original composites~\ref{tab:deqa}. 
    Fine-tuning on RPHarmony proves especially beneficial for such models, effectively alleviating overfitting and improving generalization in real-world settings.
    
    \item Leveraging the rich real-world prior embedded in the LDM architecture, our model produces more visually pleasing and realistic harmonization results, with natural foreground-background consistency even in challenging scenes.
\end{itemize}

\subsection{Why We Do Not Use RealHM Ground Truth for Quantitative Evaluation}

The ground truth (i.e., manual adjustment (MA) results) in the RealHM dataset is manually annotated using Photoshop by matching foreground appearance attributes—such as color, brightness, and contrast—to the background, with further adjustments in some regions. 

However, some MA results are not consistently reliable; the foreground is overly influenced by the background, resulting in unnatural appearances. 
Accordingly, the MA results yield a lower average quality score compared to the original composite images~\ref{tab:deqa}.
Therefore, quantitative evaluation based on such ground truth is unreliable. 

\begin{figure*}
    \centering
    \includegraphics[width=1.0\linewidth]{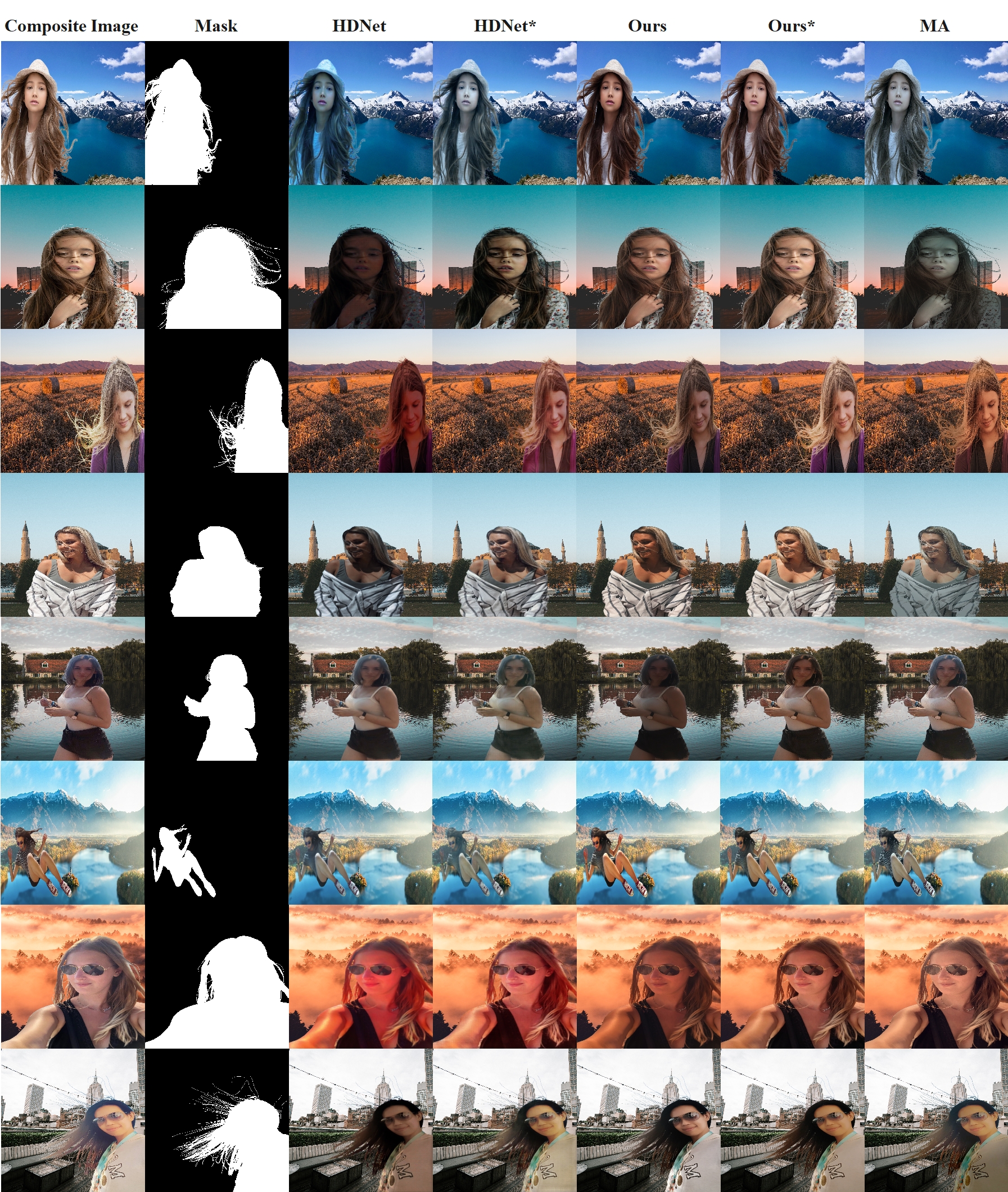}
    \caption{\textbf{Qualitative results on the RealHM dataset~\cite{jiang2021sshselfsupervisedframeworkimage}.} * denotes models that are fine-tuned on the RPHarmony dataset. ``MA'' stands for manual adjustment results, referring to the harmonized images in the RealHM dataset created via manual editing in Photoshop.}
    \label{fig:real-hr2}
\end{figure*}

For example, in the first and second rows of Fig.~\ref{fig:real-hr2}, the girl's face appears unnaturally blue due to the influence of the sky or lake. 
Similar artifacts are present in other images as well, reflecting an overemphasis on color matching rather than realistic lighting.
In contrast to these less reliable MA results, our model generates harmonized images that are more faithful to realistic lighting and appearance.

\begin{table}[t]
  \centering
  \begin{tabular}{ccc}
    \toprule[1.25pt]
    Index & FLOPs (G) & Params (M) \\
    \midrule
    1 & 0.42 & 1.44 \\
    2 & 0.43 & 5.74 \\
    3 & 0.46 & 22.94 \\
    4 & 0.46 & 22.94 \\
    \bottomrule[1.25pt]
  \end{tabular}
    \caption{\textbf{Computational Cost and Parameter Count of the MACA Module} The "Index" denotes the position or ID of each MACA module within the Harmony Controller. }
  \label{tab:complexity-maca}
\end{table}

\section{7\quad Computational Efficiency Analysis of the MACA Module}

Tab.~\ref{tab:complexity-maca} presents the parameter count and computational cost (measured in FLOPs) of the MACA module. The results indicate that the MACA module is computationally efficient.

\section{8\quad  Limitations and Future Work}

While our method and dataset contribute to improving the performance of image harmonization models and achieve state-of-the-art results, several challenges remain to be addressed. 
First, our model has a relatively large number of parameters, which may limit its applicability in high-resolution or real-time inference scenarios. This calls for future work on model compression and task-specific lightweight design. 
Second, the RPHarmony dataset provides more challenging image harmonization pairs. If a model performs poorly or struggles to converge, fine-tuning on RPHarmony may not lead to improved generalization on real images.

While our method is compatible with a wide range of diffusion models, our current experiments are limited to using Stable Diffusion as the base model due to computational resource constraints. In future work, we plan to explore its application to DiT-based models~\cite{DiT} to further improve performance.
Additionally, we plan to apply Random Poisson Blending to construct larger-scale datasets and investigate the relationship between dataset scale and model generalization.


\bibliography{aaai2026}